\documentclass[10pt]{article}

\usepackage[preprint]{tmlr}
\usepackage{amsmath,amssymb}
\usepackage{booktabs,longtable}
\usepackage{caption}
\usepackage{graphicx}
\usepackage{microtype}
\usepackage{fancyvrb}
\usepackage{float}
\usepackage[section]{placeins}
\usepackage{xcolor}
\usepackage[colorlinks=true,linkcolor=blue,citecolor=blue,urlcolor=blue]{hyperref}
\hypersetup{
  pdftitle={Same Quantity, Different Answer: Numerical Representation Invariance in Language Models},
  pdfauthor={Ephraim Atta-Duncan}
}

\title{Same Quantity, Different Answer:\\
Numerical Representation\\
Invariance in Language Models}

\author{Ephraim Atta-Duncan\\
\texttt{hello@ephraimduncan.com}}

\begin{document}
\maketitle

\begin{abstract}
Numerically equivalent word problems should yield the same canonical
answer whether a quantity is written as a decimal, fraction, percentage,
number word, scientific notation, or an exactly converted unit. We
generate 3{,}600 exact-rational problems and 8{,}600 prompts spanning
five identity-preserving transformation families, and evaluate five
open-weight systems. After a fixed syntax audit that normalizes common
answer forms without an LLM judge, canonical accuracy is .969--.996,
but orbit correctness falls to .848--.981 and orbit invariance to
.851--.981; invariant-but-wrong orbits account for at most .003. Most
of the broad strict-parser collapse arises because multiplication-form
scientific notation lies outside the implemented number grammar,
illustrating how evaluator interfaces can masquerade as reasoning
failures. A distinct semantic pathology remains: Mistral Small 4 scores
.699 on unit-converted inputs and produces 265 errors differing from the
label by exact powers of ten. In a separate 9{,}000-call experiment that
allocates equal calls to the compared arms, representation consensus
does not outperform paraphrase consensus on a low-error subset and
produces substantially more false alarms. The supplementary archive
contains the frozen benchmark, evaluation and audit records, consensus
raw responses, manifests, analysis code, and a one-command paper build.
\end{abstract}

\section{Introduction}

The correctness of a quantitative answer should not depend on whether a
volume is written as \texttt{3/4 L} or \texttt{0.75 L}. Numerical
\emph{representation invariance} requires the same normalized answer
under an identity-preserving rewrite of the input, including an exact
rewrite into equivalent units. ParaRel provides a methodological
precedent for checking consistency under paraphrase~\citep{elazar2021};
our transformations instead have algebraically known identity
relations. Aggregate benchmarks obscure violations because they expose
only one surface form per item.

Prior work establishes several neighboring effects. Language models
encode numbers imperfectly~\citep{wallace2019,thawani2021};
digit-segmentation choices can affect arithmetic
accuracy~\citep{singh2024tokenization,mcleish2024}; semantically
irrelevant prompt formatting can move task accuracy
substantially~\citep{sclar2024}; and numeral, unit, or script
changes can alter quantitative performance~\citep{numcot2024,bui2025,
reddy2026}. These results motivate an exact, per-item test that places
heterogeneous representation changes in one framework without assuming
a tokenization or decoding mechanism.

Our framework has three parts. First, each item is a typed semantic
program with an exact rational answer. A renderer expands it into an
\emph{orbit} of equivalent prompts under constrained
decimal/fraction, scientific-notation, percent/decimal, digits/words,
and input-unit transformations. Equivalence follows from the program
and registered exact conversion factors, rather than from post-hoc
judgment.

Second, we separate a narrow parser interface from an audited value
layer. This distinction is necessary because the natural-language
instruction uses the placeholder \texttt{<number>} but does not enumerate
the parser's accepted syntaxes. The strict layer therefore measures
compatibility with the implemented parser. The audit strips markdown
emphasis, case-folds the answer tag, and normalizes multiplication-form
scientific notation to e-notation before re-running that same parser.
It is deliberately narrower than a general answer extractor.

Third, the orbit supplies a ground-truth-free disagreement signal at
inference time, in the spirit of metamorphic testing and selective
prediction. We compare representation consensus with equal-call
self-consistency and fixed-paraphrase controls under matched sampling
settings.

Running five pinned systems---Qwen3.5 at 4B and 9B, the same 9B
checkpoint at Q4\_K\_M and Q8\_0, GPT-OSS-20B, and Mistral Small
4---produces four findings. First, strict-parser compatibility is highly
surface-sensitive, chiefly because standard multiplication-form
scientific notation lies outside the implemented grammar. Second,
audited semantic outcomes remain representation-sensitive: audited
orbit correctness is .848--.981 and audited invariance is .851--.981,
despite canonical correctness of .969--.996. Third, Mistral's
unit-conversion weakness is not repaired by the audit and is concentrated
in two eligible templates. Fourth, H3 is unsupported on the evaluated
120-base subset: representation and paraphrase AURC are indistinguishable
within paired-bootstrap uncertainty, while representation views generate
many more false alarms.

\paragraph{Contributions.}
\begin{enumerate}
  \item An exact-rational orbit framework that reports correctness,
    invariance, and consistent-wrong outcomes per base problem.
  \item A frozen benchmark of 3{,}600 typed base problems and 8{,}600
    prompts, with explicit eligibility masks and 1{,}000 pairs per
    transformation family.
  \item A transparent separation between strict-parser compatibility and
    a fixed audited value layer, including paired confidence intervals,
    directional discordance, multiplicity-adjusted tests, and audited
    orbit metrics.
  \item A five-system descriptive study with a same-checkpoint Q4/Q8
    comparison and template-stratified analysis of the largest value
    failure.
  \item A secondary equal-call consensus analysis with paired uncertainty
    and fallback-order sensitivity, reported as an unsupported H3 rather
    than evidence of equivalence.
\end{enumerate}

\section{Related Work}

\paragraph{Numeracy and number encoding.}
\citet{wallace2019} showed that static and contextual embeddings encode
numeracy imperfectly, and \citet{thawani2021} survey consequences of
number surface forms in NLP. \citet{singh2024tokenization} find that
digit-segmentation conventions affect arithmetic accuracy, while
\citet{mcleish2024} improve arithmetic with digit-position embeddings.
These works establish plausible mechanisms for other settings. Our
experiment does not manipulate tokenizers, so its cross-system patterns
cannot identify tokenization as the cause.

\paragraph{Perturbing math word problems.}
GSM-Symbolic~\citep{mirzadeh2025gsmsymbolic} templates GSM8K items and
varies names and numeric values, demonstrating instability across
instantiations; the values change but their representations do not, and
the analysis is predominantly aggregate over 50-instantiation sets.
Numeric-remapping attacks~\citep{remap2026} remap values automatically
and recompute labels via LLM-inferred programs, with unspecified
numeric precision. Both lines vary the instantiated numerical
quantities rather than their equivalent surface representations. We
hold the underlying quantities and requested answer fixed and vary only
how the input quantities are written, with labels that are exact by
construction rather than recomputed.

\paragraph{Representation and format sensitivity.}
NUMCoT~\citep{numcot2024} studies numeral and measurement-unit choices
in chain-of-thought reasoning and evaluates them with task-specific
answer extraction. \citet{bui2025} vary measurement systems over
compiled real-world facts and report increased test-time compute for
underrepresented systems. \citet{sclar2024} document large accuracy
changes under prompt-format variations that preserve content.
\citet{reddy2026} provide the nearest numeral-format comparison: they
hold arithmetic fixed while varying scripts and formatting conventions
on expressions, and distinguish correct-value/wrong-format responses.
Our narrower contribution is exact-rational generation and unit
canonicalization across five jointly analyzed families, with both orbit
correctness and invariance measured per item.

\paragraph{Behavioral and metamorphic testing.}
CheckList organizes behavioral tests as minimum-functionality,
invariance, and directional-expectation tests~\citep{ribeiro2020checklist}.
METAL applies metamorphic relations to analyze LLM
qualities~\citep{hyun2023metal}, and LLMorph develops natural-language
metamorphic transformations for NLP tasks~\citep{cho2025llmorph}. Our
orbits instantiate the same test-oracle idea with typed quantitative
programs: the input relation and exact output relation are generated
together, and all answers are canonicalized to rational values.

\paragraph{Consistency, selective prediction, and consensus.}
ParaRel~\citep{elazar2021} measures factual consistency under
paraphrase and documents substantial inconsistency. SelfCheckGPT uses
cross-sample agreement as a hallucination signal~\citep{manakul2023};
selective prediction~\citep{elyaniv2010selective,geifman2017selective}
and selective question answering~\citep{kamath2020} formalize the
risk--coverage tradeoff. Self-consistency prompting aggregates
resampled chains~\citep{wang2023selfconsistency}. Our H3 experiment
isolates a different source of diversity---equivalent numerical
representations---against equal-call resampling and paraphrase controls.
The logic follows metamorphic testing~\citep{chen1998metamorphic}, but
the empirical question is whether dissent predicts errors rather than
merely surface sensitivity.

\section{Framework and Benchmark}

\subsection{Semantic programs, orbits, and the two properties}
\label{sec:framework}

The statistical unit is a \textbf{base problem}
$b=(P,\theta,d,u^\ast,y)$: a typed program $P$ with rational parameters
$\theta$, a physical dimension $d$, a canonical unit $u^\ast$, and an
exact answer $y\in\mathbb{Q}$. A renderer $\rho_g$ maps $b$ to a prompt
under each applicable view $g\in G_b$. The \textbf{orbit} is
$\{\rho_g(b):g\in G_b\}$.

Let $C$ be a parser and exact unit canonicalizer, and let
$z_{b,g}=C(f(\rho_g(b)))$, with $z_{b,g}$ undefined when parsing fails.
\emph{Orbit correctness} requires
\[
  z_{b,g}=y\qquad\text{for every }g\in G_b,
\]
whereas \emph{orbit invariance} requires every $z_{b,g}$ to be defined
and
\[
  z_{b,g}=z_{b,h}\qquad\text{for every }g,h\in G_b.
\]
An orbit may therefore be invariant but consistently wrong. We report
that event separately rather than treating agreement as correctness.

We instantiate $C$ twice: the strict parser $C_s$ and the audited
canonicalizer $C_a$, which applies the fixed rewrite in
Section~\ref{sec:audit} before the same parser. The unit family is an
input-side invariance test: statement quantities are restated in an
exact sub-unit while the question still requests the canonical unit.
Output values in any whitelisted unit are converted by a rational
factor. Label construction, registered unit conversion, and answer
comparison use rational arithmetic; model generation itself is not
claimed to avoid floating-point computation.

\subsection{Transformation families}

\begin{table}[t]
\caption{Transformation families. Each transformed prompt preserves the
expected canonical answer on its eligible bases. Unit conversion changes
only quantities in the statement; the requested answer unit remains
canonical.}
\label{tab:families}
\centering
\small
\begin{tabular}{p{0.24\linewidth}p{0.25\linewidth}p{0.43\linewidth}}
\toprule
Family & Example & Eligibility constraint \\
\midrule
Decimal / Fraction & $0.5 \leftrightarrow 1/2$ &
Terminating decimal, denominator $2^a5^b$ \\
Scientific Notation & $8230 \leftrightarrow 8.23\times10^3$ &
Exact three-significant-digit integer and supported exponent \\
Percentage / Decimal & $25\% \leftrightarrow 0.25$ &
Dimensionless proportion in $(0,1)$ \\
Digits / Number Words & $12 \leftrightarrow$ twelve &
Integer multiplier in $[2,12]$ \\
Unit Conversion & 2\,m $\leftrightarrow$ 200\,cm &
Registered exact sub-unit for the physical dimension \\
\bottomrule
\end{tabular}
\end{table}

The constraints in Table~\ref{tab:families} make equivalence exact. No
decimal/fraction pair is rounded, and every unit factor is registered
as a rational. Scientific notation is restricted so its displayed
mantissa and exponent equal the source integer exactly. Applicability
is part of the estimand: per-view and paired results include only bases
for which that family was generated.

\subsection{Dataset construction}

\begin{table}[t]
\caption{Template allocation. Counts in parentheses are transformed
prompts; every base also has one canonical prompt. The allocations sum
to 1{,}000 prompts for each transformation family.}
\label{tab:templates}
\centering
\small
\begin{tabular}{lrl}
\toprule
Template & Bases & Additional rendered views \\
\midrule
Combined Volume & 200 & Fraction (200), Unit (200) \\
Difference Distance & 200 & Fraction (200), Unit (200) \\
Repeated Length & 1{,}000 & Fraction (200), Words (1{,}000), Unit (300) \\
Fraction of Capacity & 1{,}000 & Fraction (200), Percent (1{,}000), Unit (300) \\
Rate Speed & 200 & Fraction (200) \\
Large-Volume Addition & 1{,}000 & Scientific (1{,}000) \\
\bottomrule
\end{tabular}
\end{table}

The six templates yield 3{,}600 bases and 8{,}600 prompts
(Table~\ref{tab:templates}). Each template uses a private pseudorandom
stream under seed 20260725; duplicate argument tuples are rejected.
Eligibility masks use fixed, template-stratified quotas. Consequently,
a pooled family gap is descriptive over that family's eligible
template mixture, not a causal estimate separable from operation,
template, or exponent range. Scientific notation, percentage, and
number words each occur in only one template. We therefore scope claims
to the generated support and stratify the largest unit result below.

Generator version 0.2.0 and prompt version p1 produce
\texttt{base\_benchmark.jsonl} and
\texttt{views\_benchmark.jsonl}. SHA-256 identifiers and generation
parameters are included in the supplement; regeneration is
byte-identical, and expected answers round-trip through the public
parser.

\subsection{Prompt, strict parser compatibility, and audited values}
\label{sec:contract}
\label{sec:audit}

Every prompt says: \emph{``Return exactly one final line in this form:
\texttt{FINAL: <number> <unit>}''}. The placeholder
\texttt{<number>} does not enumerate the implemented grammar. The strict
parser accepts signed integers, finite decimals with optional comma
grouping, integer fractions, and e-notation, but not the conventional
multiplication form $m\times10^k$. It also accepts the last of multiple
\texttt{FINAL} lines while flagging the response. Strict scores are
therefore named \emph{strict-parser compatibility}; calling every
rejection a model contract violation would overstate what the prompt
specified.

The parser canonicalizes whitelisted units with rational factors.
Failures have fixed precedence:
\texttt{truncation\_error} (finish reason other than stop),
\texttt{format\_error} (no uppercase \texttt{FINAL:} line),
\texttt{parse\_error}, \texttt{unit\_error}, then
\texttt{value\_error}. There is no LLM judge or free-form repair. The
operational grammar summary appears in Appendix~\ref{app:parser}; the
included parser source is authoritative.

Audit version \texttt{audit\_v2} is applied to strict-failed responses:
remove markdown emphasis, case-fold \texttt{FINAL:}, and normalize
$m\times10^k$, \texttt{x10\^{}k}, and Unicode-superscript scientific
forms to e-notation. The unchanged parser then runs again. A recovered
item shows that the transcript contains the expected value under this
narrow normalization; it does not prove a particular internal
calculation. Residual parseable wrong values are tagged
\emph{magnitude-drop} when
$\mathrm{parsed}\times10^k=\mathrm{expected}$ for an integer
$0<|k|\leq15$. All comparisons are exact.

\subsection{Metrics, hypotheses, and statistics}
\label{sec:metrics}

For both strict and audited canonicalizers we report per-view
correctness, orbit correctness, orbit invariance, and
consistent-wrong rate. Per-view denominators are the eligible bases,
not all 3{,}600 bases. The orbit gap is canonical correctness minus
orbit correctness. For each scoring layer, family tables report
canonical-minus-transformed correctness on the 1{,}000 paired bases,
directional discordance, and exact McNemar tests.

The protocol fixed a 3-percentage-point practical-effect threshold:
30 outcomes per 1{,}000-pair family. This is a pragmatic materiality
choice, not a power-derived boundary. Holm adjustment is performed
separately within each scoring layer and system over exactly five family
tests; it does not cover the five systems, orbit summaries, or
exploratory strata. Family-gap and orbit-gap intervals use 10{,}000
base-clustered bootstrap draws. Audited orbit rates use two-sided 95\%
Wilson intervals. The project plan fixed these choices before
cross-model runs but was not deposited in an external registry, so we
use \emph{protocol-fixed}, not \emph{preregistered}.

Operational H1 asks whether any strict-parser family gap is at least
three points and Holm-significant; operational H2 asks whether the
strict-parser orbit gap is at least three points. Because of the
prompt/parser mismatch, these hypotheses test the implemented
interface, not arithmetic invariance alone. We report the audited paired
analysis as the primary semantic result without relabeling the original
hypotheses. H3 asks whether representation consensus improves
risk--coverage over both equal-call controls. AURC averages uniformly
over orderings within equal-agreement ties.

\section{Experimental Setup}

\subsection{Systems}
\label{sec:systems}

\begin{table}[H]
\caption{Evaluated systems. Each checkpoint, quantization, and decoding
combination is a distinct system. Qwen3.5-9B Q4/Q8 shares a source
checkpoint; GPT-OSS necessarily uses its reasoning mode and a larger
generation cap.}
\label{tab:systems}
\centering
\small
\setlength{\tabcolsep}{4pt}
\begin{tabular}{lllrr}
\toprule
System & Architecture & Quantization & Reasoning & Cap \\
\midrule
Qwen3.5-4B & dense, 4B & Q4\_K\_M & disabled & 1024 \\
Qwen3.5-9B & dense, 9B & Q4\_K\_M & disabled & 1024 \\
Qwen3.5-9B & dense, 9B & Q8\_0 & disabled & 1024 \\
GPT-OSS-20B & MoE, 3.6B active & MXFP4 & low & 4096 \\
Mistral Small 4 & MoE, 119B/6.5B active & Q4\_K\_M & disabled & 1024 \\
\bottomrule
\end{tabular}
\end{table}

We use the official Qwen3.5-4B and Qwen3.5-9B
releases~\citep{qwen35_4b_card,qwen35_9b_card},
GPT-OSS~\citep{openai2025gptoss}, and Mistral Small
4~\citep{mistral2026small4}. Repository revisions,
source revisions when recorded, quantized filenames, and SHA-256 hashes
are in the stable provenance manifest. Qwen and Mistral use their
documented chat-template controls to disable reasoning; GPT-OSS runs at
low reasoning effort.

\subsection{Inference protocol}

All primary runs use one pinned \texttt{llama.cpp}
build~\citep{gerganov2023llamacpp} (b10107, commit c0bc8591e; CUDA
backend) on NVIDIA H100 80\,GB HBM3 GPUs, with one request in flight.
Decoding is greedy (temperature 0, top-$k$ 1, top-$p$ 1, seed
20260725), but we do not infer mathematical determinism from those
settings. A ten-prompt repeat check was 10/10 byte- and answer-identical
for Qwen3.5-4B, both Qwen3.5-9B quantizations, and GPT-OSS. Mistral was
answer-identical on 9/10; one answer changed from 33.79 to 33.7925.
These observations characterize only the repeated prompts. Every run
stores request order, sampling settings, model and runtime identifiers,
and raw responses.

\section{Results: Representation Sensitivity Across Systems}

\begin{figure}[!ht]
\centering
\includegraphics[width=0.86\linewidth]{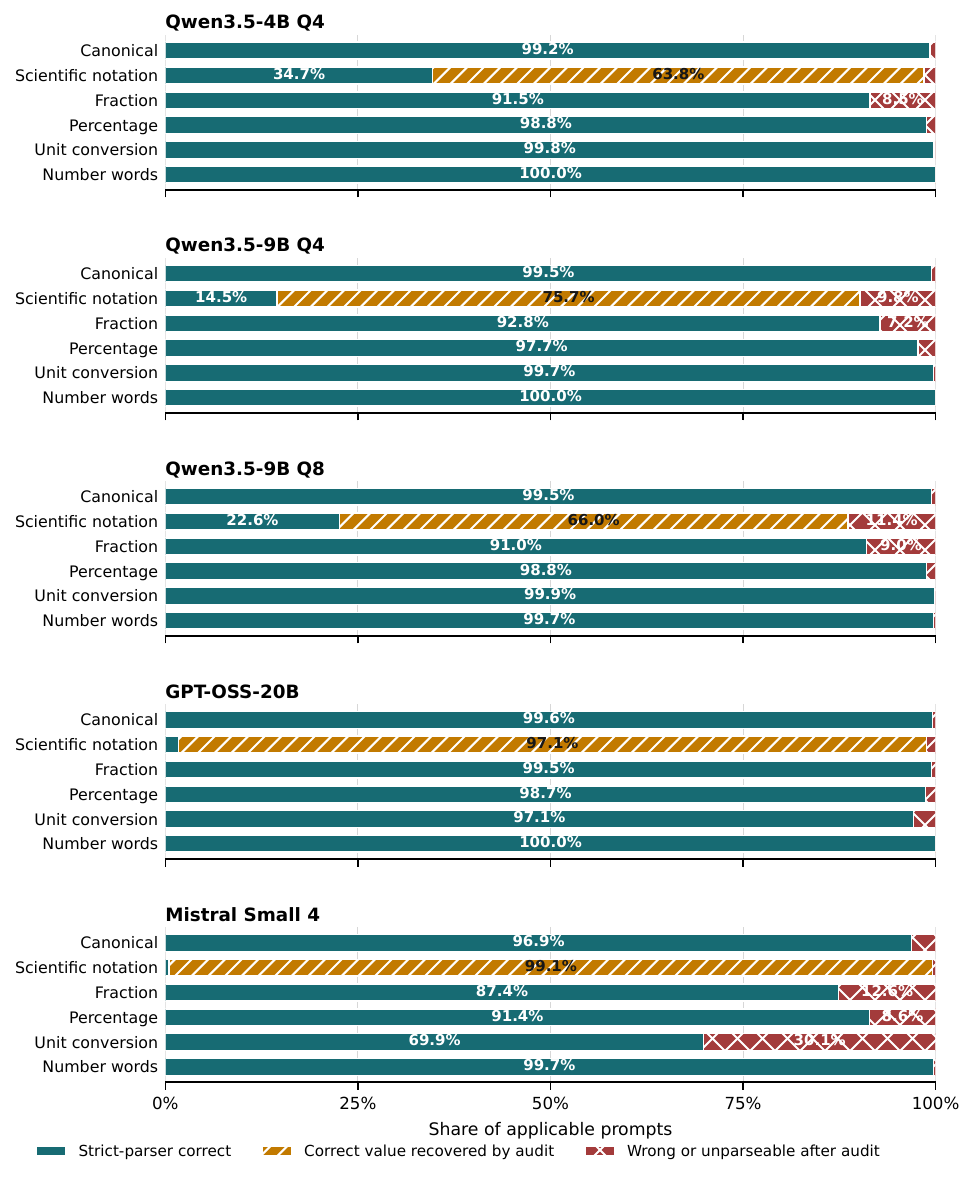}
\caption{Outcome composition on eligible prompts:
$n{=}3{,}600$ for canonical and $n{=}1{,}000$ for each transformed
view. Solid teal is strict-parser correct; diagonally hatched amber is
correct after the fixed audit; cross-hatched red is wrong or
unparseable after audit. Percent labels use one decimal place, and
machine-readable counts are provided with the figure. Scientific
notation is dominated by audit recoveries, whereas Mistral unit
conversion contains a large residual-failure segment.}
\label{fig:outcomes}
\end{figure}
\FloatBarrier

Figure~\ref{fig:outcomes} separates parser-interface mismatch from
residual value failure. Scientific notation produces broad audit
recovery across all systems. Fraction inputs show smaller residual
losses for Qwen and Mistral, and Mistral unit conversion forms a
distinct value-failure pattern. A canonical-only evaluation would hide
all three structures.

\subsection{Audited semantic paired results}

Under the implemented strict parser, operational H1 and H2 hold for all
five systems. We retain those protocol-fixed interface results in
Appendix~\ref{app:full}; the main analysis instead foregrounds audited
values because multiplication-form scientific notation is outside the
strict grammar.

\begingroup
\footnotesize
\setlength{\tabcolsep}{3pt}
\begin{longtable}{@{}llrlll@{}}
\caption{Audited paired results, $n{=}1{,}000$ per family. C/T is
canonical/transformed correctness; the gap is C minus T with a 95\%
base-bootstrap interval; c/t is canonical-only/transformed-only
discordance. Holm adjustment is within each system's five audited tests.
Bold gaps exceed three points and have adjusted $p<.05$; this descriptive
audit is not a relabeling of H1.}
\label{tab:audited}\\
\toprule
System & Family & C/T & Gap [95\% CI] & $p_{\text{Holm}}$ & c/t \\
\midrule
\endfirsthead
\caption[]{Audited paired results (continued).}\\
\toprule
System & Family & C/T & Gap [95\% CI] & $p_{\text{Holm}}$ & c/t \\
\midrule
\endhead
\bottomrule
\endlastfoot
Qwen3.5-4B Q4 & Sci.\ notation & 1.000/.985 & $+.015$ [.008,.023] & .00024 & 15/0 \\
 & Unit conversion & .995/.998 & $-.003$ [$-$.008,.002] & 1.000 & 2/5 \\
 & Fraction & .987/.915 & $\mathbf{+.072}$ [.054,.091] & $1.7\times10^{-14}$ & 82/10 \\
 & Percentage & .984/.988 & $-.004$ [$-$.013,.005] & 1.000 & 9/13 \\
 & Number words & 1.000/1.000 & $.000$ [$-$.003,.003] & 1.000 & 0/0 \\
\addlinespace
Qwen3.5-9B Q4 & Sci.\ notation & .998/.902 & $\mathbf{+.096}$ [.078,.115] & $3.1\times10^{-27}$ & 97/1 \\
 & Unit conversion & .993/.997 & $-.004$ [$-$.010,.001] & .578 & 2/6 \\
 & Fraction & .989/.928 & $\mathbf{+.061}$ [.044,.078] & $3.1\times10^{-12}$ & 70/9 \\
 & Percentage & .991/.977 & $+.014$ [.004,.025] & .038 & 21/7 \\
 & Number words & 1.000/1.000 & $.000$ [$-$.003,.003] & 1.000 & 0/0 \\
\addlinespace
Qwen3.5-9B Q8 & Sci.\ notation & .999/.886 & $\mathbf{+.113}$ [.093,.133] & $2.8\times10^{-32}$ & 114/1 \\
 & Unit conversion & .997/.999 & $-.002$ [$-$.006,.002] & .909 & 1/3 \\
 & Fraction & .990/.910 & $\mathbf{+.080}$ [.062,.098] & $4.5\times10^{-18}$ & 87/7 \\
 & Percentage & .992/.988 & $+.004$ [$-$.004,.012] & .909 & 10/6 \\
 & Number words & 1.000/.997 & $+.003$ [.000,.007] & .750 & 3/0 \\
\addlinespace
GPT-OSS-20B & Sci.\ notation & .997/.988 & $+.009$ [.002,.017] & .141 & 12/3 \\
 & Unit conversion & .997/.971 & $+.026$ [.016,.037] & $4.3\times10^{-6}$ & 28/2 \\
 & Fraction & .993/.995 & $-.002$ [$-$.009,.005] & 1.000 & 5/7 \\
 & Percentage & .993/.987 & $+.006$ [$-$.002,.014] & .630 & 11/5 \\
 & Number words & 1.000/1.000 & $.000$ [$-$.003,.003] & 1.000 & 0/0 \\
\addlinespace
Mistral Small 4 & Sci.\ notation & .996/.996 & $.000$ [$-$.005,.006] & 1.000 & 4/4 \\
 & Unit conversion & .980/.699 & $\mathbf{+.281}$ [.252,.309] & $5.8\times10^{-75}$ & 288/7 \\
 & Fraction & .967/.874 & $\mathbf{+.093}$ [.070,.116] & $8.2\times10^{-15}$ & 119/26 \\
 & Percentage & .916/.914 & $+.002$ [$-$.016,.020] & 1.000 & 44/42 \\
 & Number words & .994/.997 & $-.003$ [$-$.007,.000] & .750 & 0/3 \\
\end{longtable}
\endgroup

Seven system--family cells exceed both the descriptive three-point
threshold and the adjusted significance criterion. The largest audited
gap is Mistral unit conversion at 28.1 points. Qwen fraction gaps are
6.1--8.0 points, and the two 9B scientific gaps are 9.6 and 11.3 points;
Section~\ref{sec:pathologies} separates cap-sensitive records from
parsed wrong values.

\begin{table}[H]
\caption{Audited orbit metrics on 3{,}600 bases. C = correctness, I =
invariance, CW = invariant but consistently wrong, and gap = canonical
correctness minus orbit correctness. Rate intervals are two-sided 95\%
Wilson intervals; gap intervals use the paired base bootstrap.}
\label{tab:orbit}
\centering
\footnotesize
\setlength{\tabcolsep}{2.5pt}
\begin{tabular}{@{}lrrrrr@{}}
\toprule
System & Canon.\ C & Orbit C [95\% CI] & Orbit I [95\% CI] & CW [95\% CI] & Gap [95\% CI] \\
\midrule
Qwen3.5-4B Q4 & .9925 & .9625 [.9558,.9682] & .9633 [.9567,.9690] & .0008 [.0003,.0024] & .0300 [.0244,.0358] \\
Qwen3.5-9B Q4 & .9947 & .9422 [.9341,.9494] & .9422 [.9341,.9494] & .0000 [.0000,.0011] & .0525 [.0453,.0600] \\
Qwen3.5-9B Q8 & .9947 & .9350 [.9265,.9426] & .9350 [.9265,.9426] & .0000 [.0000,.0011] & .0597 [.0519,.0675] \\
GPT-OSS-20B & .9958 & .9806 [.9755,.9846] & .9808 [.9758,.9848] & .0003 [.0000,.0016] & .0153 [.0114,.0194] \\
Mistral Small 4 & .9689 & .8481 [.8360,.8594] & .8508 [.8388,.8621] & .0028 [.0015,.0051] & .1208 [.1103,.1319] \\
\bottomrule
\end{tabular}
\end{table}

Canonical-minus-audited-orbit gaps are 3.0, 5.3, 6.0, 1.5, and
12.1 points in table order. Thus the same three-point descriptive
criterion is met by four systems, not GPT-OSS. Audited invariance closely
tracks audited correctness because consistent wrong orbits are rare, but
none of these estimates is perfect invariance.

\subsection{Two layers dissociate: shared audit recovery and one value pathology}
\label{sec:pathologies}

\begin{table}[H]
\caption{Scientific-notation responses under strict and audited scoring
($n{=}1{,}000$). Audit recovered counts expected values exposed by the
fixed rewrite; residual magnitude errors differ from the label by an
exact power of ten.}
\label{tab:twolayer}
\centering
\small
\begin{tabular}{lrrrrrr}
\toprule
System & Strict & Audited & Audit recovered & Mag.\ errors & Other value & Unparseable \\
\midrule
Qwen3.5-4B Q4 & .347 & .985 & 638 & 2 & 0 & 13 \\
Qwen3.5-9B Q4 & .145 & .902 & 757 & 0 & 9 & 89 \\
Qwen3.5-9B Q8 & .226 & .886 & 660 & 0 & 4 & 110 \\
GPT-OSS-20B & .017 & .988 & 971 & 0 & 12 & 0 \\
Mistral Small 4 & .005 & .996 & 991 & 0 & 4 & 0 \\
\bottomrule
\end{tabular}
\end{table}

\paragraph{Scientific notation: values mostly survive a parser mismatch.}
Strict scientific correctness ranges from .005 to .347, while audited
correctness ranges from .886 to .996
(Table~\ref{tab:twolayer}). The rewrite recovers 638--991 expected
values per system from forms such as
\texttt{1.494 x 10\^{}4 liters}. Qwen scientific responses include 13
length finishes for 4B Q4, 85 for 9B Q4, and 110 for 9B Q8. Excluding
those records in a descriptive sensitivity analysis leaves audited
accuracy of .998 (985/987), .986 (902/915), and .996 (886/890),
respectively. This conditioning does not estimate what a higher-cap
rerun would produce. The broad strict collapse establishes
incompatibility with the implemented parser, not inability to perform
the arithmetic.

\paragraph{Unit conversion: a silent power-of-ten pattern.}
Mistral's unit view remains .699 after audit. Of 301 residual errors,
265 have an exact power-of-ten relation to the label: 96 by $10$, 70 by
$10^2$, and 99 by $10^3$. This pattern is consistent with carrying a
sub-unit count into a canonical-unit answer, but output traces do not
identify the model's internal mechanism. A syntax-only constraint would
not catch a parseable wrong value.

\begin{table}[H]
\caption{Mistral unit-view result by eligible template. The pooled .699
score combines materially different strata; no unit-conversion claim is
made outside these four template/operation combinations.}
\label{tab:unit-strata}
\centering
\small
\begin{tabular}{lrrr}
\toprule
Template & $n$ & Audited correctness & Magnitude-drop errors \\
\midrule
Combined Volume & 200 & .980 & 0 \\
Difference Distance & 200 & .935 & 12 \\
Repeated Length & 300 & .443 & 158 \\
Fraction of Capacity & 300 & .610 & 95 \\
\bottomrule
\end{tabular}
\end{table}

\begin{figure}[!t]
\begin{Verbatim}[frame=single,fontsize=\scriptsize]
PROMPT (unit view; Repeated Length item 383)
  Solve the quantitative problem. Return exactly one final line in this form:
  FINAL: <number> <unit>

  A carpenter cuts 11 boards. Each board is 330 centimeters long.
  What is the total length of all the boards in meters?

RESPONSE (Mistral Small 4, greedy)
  CALC: 11 * 330 = 3630 centimeters

  FINAL: 3630 meters

EXPECTED
  FINAL: 36.3 meters
\end{Verbatim}
\caption{Illustrative transcript excerpt. Indentation is added for
layout; prompt and response text are otherwise reproduced from the raw
record, which is authoritative. The parseable answer relabels a
centimeter count as meters, so exact comparison to the semantic label
is required.}
\label{fig:transcript}
\end{figure}
\FloatBarrier

\subsection{Audited value correctness by view}

\begin{table}[H]
\caption{Audited value correctness on each eligible view. Scientific
notation largely recovers after the fixed normalization; Mistral unit
conversion does not.}
\label{tab:value}
\centering
\small
\begin{tabular}{lrrrrrr}
\toprule
System & Canonical & Sci.\ notation & Unit conv. & Fraction & Percentage & Words \\
\midrule
Qwen3.5-4B Q4 & .9925 & .985 & .998 & .915 & .988 & 1.000 \\
Qwen3.5-9B Q4 & .9947 & .902 & .997 & .928 & .977 & 1.000 \\
Qwen3.5-9B Q8 & .9947 & .886 & .999 & .910 & .988 & .997 \\
GPT-OSS-20B & .9958 & .988 & .971 & .995 & .987 & 1.000 \\
Mistral Small 4 & .9689 & .996 & \textbf{.699} & .874 & .914 & .997 \\
\bottomrule
\end{tabular}
\end{table}

The audit changes the scientific interpretation but not every family.
Audited fraction correctness is .915 for Qwen3.5-4B Q4, .928 for
Qwen3.5-9B Q4, .910 for Qwen3.5-9B Q8, .995 for GPT-OSS, and .874 for
Mistral. Words remain .997--1.000. These are within-eligibility
descriptions, not family effects averaged over a common base set.

\subsection{Observed scale and quantization comparisons}
\label{sec:scale}

Within the evaluated Qwen releases, moving from 4B to 9B under
Q4\_K\_M changes strict scientific correctness from .347 to .145 while
canonical correctness rises. For the same 9B source checkpoint, Q8\_0
changes strict scientific correctness to .226 and audited scientific
correctness from .902 to .886; audited fraction correctness changes
from .928 to .910. These two controlled comparisons show that the
observed sensitivity is present at both tested precisions. They do not
establish a general relationship between scale, quantization, and
invariance: there is one checkpoint pair, no unquantized baseline, and
no replication across model families.

\subsection{Completion length and truncation}
\label{sec:cost}

\begin{table}[H]
\caption{Completion-token median [Q1, Q3] and total length-finished
responses. Quartiles use the inclusive empirical definition. Values are
comparable across views within a system; tokenizer and chat-template
differences preclude a controlled cross-system token-cost comparison.}
\label{tab:cost}
\centering
\footnotesize
\setlength{\tabcolsep}{3pt}
\begin{tabular}{lrrrrr}
\toprule
System & Canonical & Fraction & Sci.\ notation & Unit & Length-finished \\
\midrule
Qwen3.5-4B Q4 & 113 [103,137] & 272 [225.8,343] & 275 [261,327] & 168 [153,198] & 22 \\
Qwen3.5-9B Q4 & 189 [128,257] & 302 [260,368] & 467 [371,610.3] & 197 [169,211] & 111 \\
Qwen3.5-9B Q8 & 190 [126,263.3] & 316.5 [269,381] & 520.5 [466.5,650.3] & 205 [180,222] & 144 \\
GPT-OSS-20B & 32 [28,34] & 68 [49,96] & 47 [37,55] & 53 [46,60] & 0 \\
Mistral Small 4 & 12 [11,12] & 109 [60,178] & 18 [17,18] & 11 [10,13] & 0 \\
\bottomrule
\end{tabular}
\end{table}

Within each Qwen run, scientific prompts have longer median completions
than canonical prompts, and length finishes concentrate in the
scientific view (13, 85, and 110 respectively). Fraction prompts also
lengthen completions for all five systems. These are descriptive
associations under greedy decoding; repeat checks do not establish that
the additional tokens are deterministic, nor that length alone causes
the accuracy pattern.
\section{Secondary Analysis: Representation Consensus (H3)}
\label{sec:consensus}

\paragraph{Design and scope.}
The subset contains 120 bases: the first 30 by identifier from Combined
Volume, Difference Distance, Repeated Length, and Fraction of Capacity.
It excludes Rate Speed (too few applicable views) and Large-Volume
Addition (only canonical and scientific views). Thus H3 does not test
the largest parser-interface failure and operates near the accuracy
ceiling.

Each arm uses three calls. Self-consistency resamples the canonical
prompt. Paraphrase consensus uses the canonical wording as position 1
and two fixed, manually authored wordings as positions 2--3; all numeric
literals and units remain unchanged. Representation consensus uses
canonical, fraction, and unit views, with percent replacing unit for
Fraction of Capacity. The literal paraphrase templates are included in
the supplement. Calls use matched seed positions and caps. The
stochastic tier evaluates all arms; the artifact key
\texttt{deterministic} denotes greedy decoding and contains the two
prompt-diverse arms.

At least two equal canonicalized values produce a prediction; otherwise
the method abstains. Appendix~\ref{app:parser} specifies the
no-abstention fallback and call-priority sensitivity analysis.

\begin{figure}[t]
\centering
\includegraphics[width=0.92\linewidth]{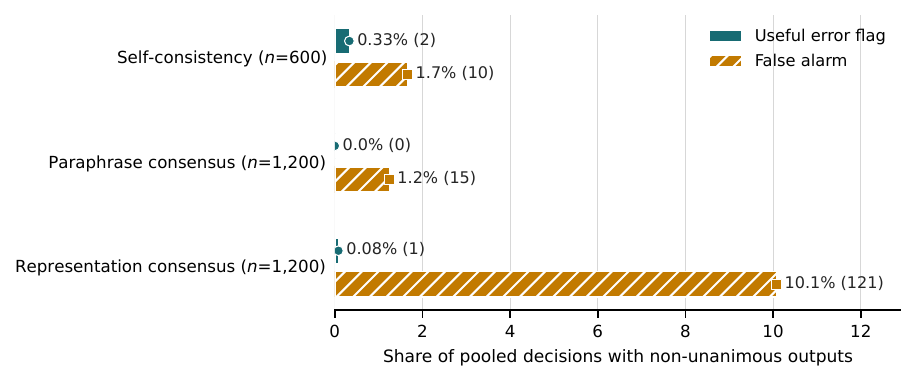}
\caption{Non-unanimous outcomes pooled over five systems. Solid teal
circles mark useful error flags; hatched amber squares mark false
alarms. Denominators are 600 for stochastic self-consistency and
1{,}200 for each prompt-diverse arm across both tiers. Representation
produces 122 flags: one useful flag (0.08\%) and 121 false alarms
(10.1\%) under the default fallback.}
\label{fig:consensus-dissent}
\end{figure}

\paragraph{H3 is unsupported and the comparison is ceiling-limited.}
Paraphrase AURC is zero in all ten system-by-tier cells.
Representation-minus-paraphrase AURC is exactly zero with paired
bootstrap interval $[0,0]$ in nine cells. In the remaining Mistral
stochastic cell, the point difference is .000069 with 95\% paired
base-bootstrap interval $[0,.000419]$. The interval includes zero, so
no cell supports a representation advantage or a resolved disadvantage.
Selective risk is zero in every paraphrase and representation cell.
These degenerate outcomes provide little power to rank the two methods;
they are evidence that H3 is unsupported on this subset, not evidence
that the methods are equivalent.

At the 3-of-3 operating point, every unanimous decision is correct.
Representation creates 122 non-unanimous decisions versus 15 for
paraphrase. A fallback-order sensitivity analysis
(Appendix~\ref{app:parser}) changes at most one useful representation
flag and does not alter the false-alarm pattern. The result is
operationally cautionary but strictly scoped to these four templates
and observed error rates.

\section{Discussion}
\label{sec:discussion}

\paragraph{The prompt/parser boundary is a measurement choice.}
The prompt's generic \texttt{<number>} placeholder does not specify that
only e-notation is accepted. Multiplication-form scientific answers are
therefore better described as incompatible with the implemented parser
than as unambiguous instruction violations. Reporting the strict and
audited layers together exposes this boundary instead of attributing
the entire strict collapse to numerical reasoning.

\paragraph{Parser compatibility does not imply numerical safety.}
Mistral's unit pattern is the converse: the final line is syntactically
accepted but numerically wrong. A grammar or JSON schema that constrains
only the output shape could accept \texttt{FINAL: 3630 meters}; semantic
comparison is still required. We do not claim every schema would accept
it, because a schema could include problem-specific numeric validation.

\paragraph{Single-surface correctness is not orbit invariance.}
After audit, orbit correctness still ranges from .848 to .981 and
invariance from .851 to .981. The strict parser exaggerates the gaps,
but a canonical-only score still misses residual representation
sensitivity. The Qwen comparisons show non-monotonic outcomes in these
specific checkpoints; they do not identify scale or quantization as a
general cause.

\paragraph{Disagreement is not automatically confidence.}
Representation views disagree more often than fixed paraphrases on the
H3 subset, but almost every additional flag is a false alarm. A
deployment should estimate uncertainty on its own error distribution
and document tie-breaking; diversity alone does not validate an
abstention policy.

\section{Limitations}

The benchmark uses synthetic templates and controlled language. Exact
equivalence and labels are strengths, but transfer to naturalistic
quantitative text is unmeasured. Family eligibility is not crossed with
every template: scientific notation, percentage, and words each occur
in one template, while fraction and unit allocations differ by
template. Pooled family results therefore mix representation with the
eligible operation and parameter distribution.

The study covers five quantized open-weight systems from three model
families on one \texttt{llama.cpp}/H100 runtime. The Qwen Q4/Q8
comparison has no unquantized baseline. Primary runs use one greedy
request per prompt. The ten-prompt repeat check is small, and Mistral
changed one answer; whole-run repeatability is unknown. Qwen
length-finished responses also make the generation cap part of the
observed scientific result. GPT-OSS has a larger cap and an unavoidable
reasoning mode, so systems are not matched on effective compute.

The audit recognizes only a fixed set of surface rewrites. It can miss
other correct values and should not be read as a complete semantic
parser. Its Wilson intervals quantify item-sampling uncertainty within
the frozen benchmark, not uncertainty over templates or model
checkpoints. H3 excludes scientific notation and has near-zero error in
the compared arms; the resulting intervals and sensitivity analysis do
not support general claims about representation consensus.

\section{Reproducibility}

The supplementary archive is part of the submission. It
contains the two frozen benchmark JSONL files, source and tests,
per-response evaluation, audit, metrics, and analysis records for the
five full runs, the five consensus-run artifacts with raw responses,
stable system aliases, sanitized manifests, figure source data, and a
per-member SHA-256 manifest. The full-run raw response transcripts and
tokenizer maps exceed the ancillary size limit and are omitted; they
are available from the author on request. Full model repository
revisions and weight hashes are recorded in
\texttt{data/manifests/model\_hashes.json}; for the sharded Mistral
artifact, the recorded file hash covers the configured entry-point
shard and is labeled accordingly rather than implying a whole-model
hash. Raw artifacts remain immutable; regenerated metrics are derived
files.

From the archive root, \texttt{make paper} regenerates consensus
analyses and per-family reports from the frozen records and rebuilds
the paper PDF; when the omitted full-run transcripts are present, it
also regenerates evaluations, audits, and figures.
\texttt{make supplement} rebuilds the deterministic supplement ZIP and
checksum. Stable
aliases replace timestamped run-directory names in the archive, and
absolute workstation paths are removed.

\section{Conclusion}

Exact orbits make numerical representation invariance measurable, but
the measurement boundary matters. The implemented strict parser is
substantially less tolerant than the prompt makes explicit, so its
.574--.785 orbit correctness is an interface diagnostic rather than a
pure arithmetic score. Under the fixed audit, orbit correctness is
.848--.981 and invariance is .851--.981, with few consistently wrong
orbits. The remaining failures are still consequential: Mistral
produces 265 exact power-of-ten unit errors concentrated in two
templates. Representation diversity also fails to improve selective
prediction on the ceiling-limited H3 subset and mostly generates false
alarms. Robust evaluation should therefore report correctness and
invariance, distinguish parser rejection from wrong values, stratify by
eligible task support, and attach uncertainty to any disagreement-based
safeguard.

\begingroup
\interlinepenalty=10000
\bibliographystyle{tmlr}
\bibliography{references}
\endgroup

\appendix

\section{Full per-family results}
\label{app:full}

Table~\ref{tab:appendix} reports strict-parser compatibility for every
system and family: canonical and transformed correctness on 1{,}000
paired bases, the paired gap with 95\% CI, Holm-adjusted exact McNemar
$p$-value, and the canonical-only/transformed-only discordance. The
scientific rows contain 4{,}250 canonical-only discordances in total;
Section~\ref{sec:pathologies} shows that most are audit-recoverable.

\begingroup
\footnotesize
\setlength{\tabcolsep}{3pt}
\begin{longtable}{@{}llrlll@{}}
\caption{Strict-parser paired results. C/T is canonical/transformed
correctness; the gap is C minus T with a 95\% base-bootstrap interval;
c/t is canonical-only/transformed-only discordance. Holm adjustment is
within each system's five strict-parser tests.}
\label{tab:appendix}\\
\toprule
System & Family & C/T & Gap [95\% CI] & $p_{\text{Holm}}$ & c/t \\
\midrule
\endfirsthead
\caption[]{Strict-parser paired results (continued).}\\
\toprule
System & Family & C/T & Gap [95\% CI] & $p_{\text{Holm}}$ & c/t \\
\midrule
\endhead
\bottomrule
\endlastfoot
Qwen3.5-4B Q4 & Fraction & .987/.915 & $+.072$ [.054,.091] & $1.3\times10^{-14}$ & 82/10 \\
 & Sci.\ notation & 1.000/.347 & $+.653$ [.624,.682] & $2.7\times10^{-196}$ & 653/0 \\
 & Percentage & .984/.988 & $-.004$ [$-$.013,.005] & 1.000 & 9/13 \\
 & Number words & 1.000/1.000 & $.000$ [$-$.003,.003] & 1.000 & 0/0 \\
 & Unit conversion & .995/.998 & $-.003$ [$-$.008,.002] & 1.000 & 2/5 \\
\addlinespace
Qwen3.5-9B Q4 & Fraction & .989/.928 & $+.061$ [.044,.078] & $3.1\times10^{-12}$ & 70/9 \\
 & Sci.\ notation & .998/.145 & $+.853$ [.830,.874] & $1.7\times10^{-256}$ & 853/0 \\
 & Percentage & .991/.977 & $+.014$ [.004,.025] & .038 & 21/7 \\
 & Number words & 1.000/1.000 & $.000$ [$-$.003,.003] & 1.000 & 0/0 \\
 & Unit conversion & .993/.997 & $-.004$ [$-$.010,.001] & .578 & 2/6 \\
\addlinespace
Qwen3.5-9B Q8 & Fraction & .990/.910 & $+.080$ [.062,.098] & $4.5\times10^{-18}$ & 87/7 \\
 & Sci.\ notation & .999/.226 & $+.773$ [.746,.799] & $2.0\times10^{-232}$ & 773/0 \\
 & Percentage & .992/.988 & $+.004$ [$-$.004,.012] & .909 & 10/6 \\
 & Number words & 1.000/.997 & $+.003$ [.000,.007] & .750 & 3/0 \\
 & Unit conversion & .997/.999 & $-.002$ [$-$.006,.002] & .909 & 1/3 \\
\addlinespace
GPT-OSS-20B & Fraction & .993/.995 & $-.002$ [$-$.009,.005] & 1.000 & 5/7 \\
 & Sci.\ notation & .997/.017 & $+.980$ [.971,.988] & $9.8\times10^{-295}$ & 980/0 \\
 & Percentage & .993/.987 & $+.006$ [$-$.002,.014] & .630 & 11/5 \\
 & Number words & 1.000/1.000 & $.000$ [$-$.003,.003] & 1.000 & 0/0 \\
 & Unit conversion & .997/.971 & $+.026$ [.016,.037] & $3.5\times10^{-6}$ & 28/2 \\
\addlinespace
Mistral Small 4 & Fraction & .967/.874 & $+.093$ [.070,.116] & $6.1\times10^{-15}$ & 119/26 \\
Mistral Small 4 (cont.) & Sci.\ notation & .996/.005 & $+.991$ [.985,.996] & $4.8\times10^{-298}$ & 991/0 \\
 & Percentage & .916/.914 & $+.002$ [$-$.016,.020] & .914 & 44/42 \\
 & Number words & .994/.997 & $-.003$ [$-$.007,.000] & .500 & 0/3 \\
 & Unit conversion & .980/.699 & $+.281$ [.252,.310] & $4.6\times10^{-75}$ & 288/7 \\
\end{longtable}
\endgroup

\section{Parser, unit registry, and audit specification}
\label{app:parser}

\paragraph{Strict parser grammar.}
The parser scans for uppercase \texttt{FINAL:} lines and parses the last,
flagging multiplicity. Accepted numbers match signed integer, decimal
with valid comma grouping, integer fraction \texttt{a/b}, or e-notation
forms. Multiplication-form scientific notation and mixed numbers are
rejected. One trailing period is tolerated. An attached percent sign is
split and applied as $1/100$. Failure precedence is truncation, missing
tag, number parse, unit lookup, then value comparison. This is an
operational summary; the included regexes, tests, and unit registry are
the complete executable specification.

\paragraph{Unit registry.}
Canonical units and rational factors are liters/milliliters
($1/1000$), kilometers/meters ($1/1000$), meters/centimeters
($1/100$), km/h, and dimensionless decimal/fraction/percent values.
Unit views rewrite only statement quantities; the question requests the
canonical unit.

\paragraph{Audit rewrite (version 2).}
For strict-failed responses, remove \texttt{**}, \texttt{*}, and
backticks; case-fold the \texttt{FINAL} tag; and rewrite
$m\times10^k$ forms using U+00D7, \texttt{x}, caret exponents, or
Unicode superscripts to e-notation. Re-run the unchanged parser.
Residual parseable values are magnitude drops exactly when
$\mathrm{parsed}\times10^k=\mathrm{expected}$ for an integer
$0<|k|\le15$. The audit changes no raw response and is not a general
natural-language extractor.

\paragraph{Consensus protocol.}
The executable specification in \texttt{consensus.py} fixes the subset,
seeds, sampling, and call order. Paraphrase position 1 is canonical;
positions 2--3 use the two fixed template skeletons in
\texttt{render.py} while preserving numbers, units, and operations.
Representation uses canonical, fraction, and unit (percent for
Fraction of Capacity). At least two equal canonicalized values produce
a prediction; otherwise the method abstains. For no-abstention
evaluation, the fallback is the modal value, then the first parsed call
when all parsed values differ; if no call parses, the fallback is wrong.
Cyclic call-priority orders produce 1--2 useful representation flags and
120--121 false alarms, versus 0 useful paraphrase flags and 15 false
alarms under every order. Source fixes acceptance, fallback, and
bootstrap details.

\end{document}